\documentclass{article}
\usepackage{spconf,amsmath,graphicx}
\usepackage{times}
\usepackage{epsfig}
\usepackage{graphicx}
\usepackage{amsmath}
\usepackage{amssymb}
\usepackage{pifont}
\usepackage{todonotes}
\usepackage{float}
\usepackage{hyperref}
\newcommand{\cmark}{\ding{51}}%
\newcommand{\xmark}{\ding{55}}%
\NewDocumentCommand{\rot}{O{90} O{1em} m}{\makebox[#2][l]{\rotatebox{#1}{#3}}}%

\title{SynthmanticLiDAR: A Synthetic Dataset for Semantic Segmentation on LiDAR Imaging}
\name{Javier Montalvo, Pablo Carballeira, Álvaro García-Martín\thanks{$^1$This work has been supported by the Ministerio de Ciencia, Innovación y Universidades of the Spanish Government under HVD (PID2021-125051OB-I00) and SEGA-CV (TED2021-131643A-I00) projects.  \\
\indent$^2$© 2024 IEEE. Personal use of this material is permitted. Permission from IEEE must be obtained for all other uses, in any current or future media, including reprinting/republishing this material for advertising or promotional purposes, creating new collective works, for resale or redistribution to servers or lists, or reuse of any copyrighted component of this work in other works.}}
\address{Video Processing and Understanding Lab, Universidad Autónoma de Madrid, Madrid, Spain}
\begin{document}
%
\maketitle
\begin{abstract}
Semantic segmentation on LiDAR imaging is increasingly gaining attention, as it can provide useful knowledge for perception systems and potential for autonomous driving. However, collecting and labeling real LiDAR data is an expensive and time-consuming task.  While datasets such as SemanticKITTI \cite{behley_semantickitti_2019} have been manually collected and labeled, the introduction of simulation tools such as CARLA \cite{dosovitskiy_carla_2017}, has enabled the creation of synthetic datasets on demand.

In this work, we present a modified CARLA simulator designed with LiDAR semantic segmentation in mind, with new classes, more consistent object labeling with their counterparts from real datasets such as SemanticKITTI, and the possibility to adjust the object class distribution. Using this tool, we have generated SynthmanticLiDAR, a synthetic dataset for semantic segmentation on LiDAR imaging, designed to be similar to SemanticKITTI, and we evaluate its contribution to the training process of different semantic segmentation algorithms by using a naive transfer learning approach. Our results show that incorporating SynthmanticLiDAR into the training process improves the overall performance of tested algorithms, proving the usefulness of our dataset, and therefore, our adapted CARLA simulator.

The dataset and simulator are available in \url{https://github.com/vpulab/SynthmanticLiDAR}.
\end{abstract}

\begin{figure}[t]
\begin{center}
   \includegraphics[width=\linewidth]{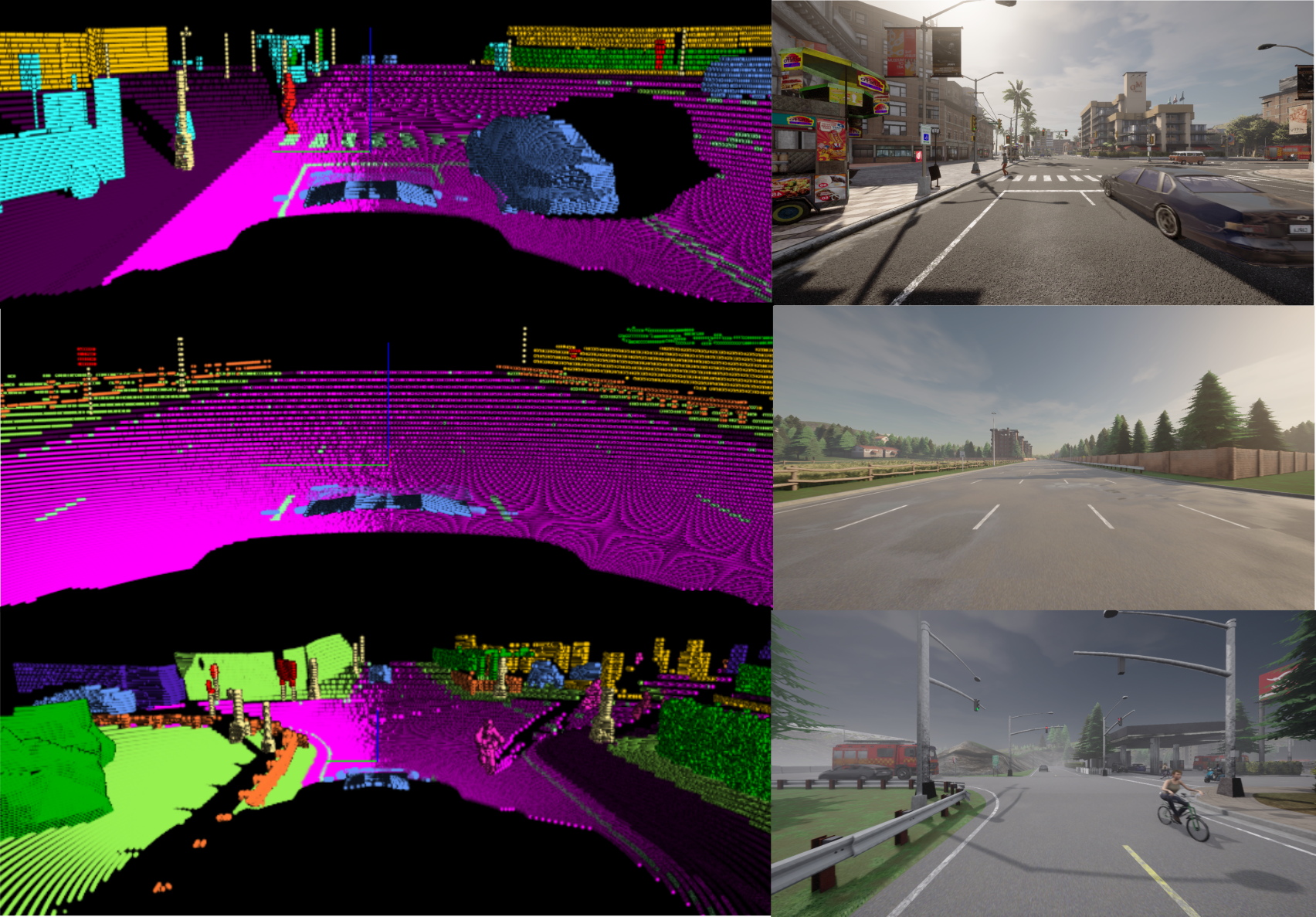}
\end{center}
   \caption{Examples of synthetic semantic LiDAR scans from SynthmanticLiDAR and an RGB image of the captured scene.}
\label{fig:examples}
\end{figure}
\section{Introduction}
As automobiles become more advanced with the integration of driving assistance systems and sensors for safety and autonomous driving purposes, novel algorithms are continually being published to extract valuable information from these sensors. In later years it has been common to equip cars with RGB cameras, to help the car stay within road lines, and with Radar sensors, to prevent frontal collisions with pedestrians or other vehicles. More recently, cars are also being equipped with LiDAR sensors, which provide a 3D representation of the surroundings of the car in the form of a point cloud, that contain precise measures for each point, and on which we can perform different computer vision tasks, such as object detection, semantic segmentation, or anomaly detection, among other methods.

Semantic segmentation for LiDAR has emerged as a critical task in the development of autonomous driving systems, as it allows the car to detect not only where things are, but also \textit{what} they are. With semantic segmentation, the vehicle can properly distinguish between things such as cars, cyclists, pedestrians, and many other types of objects, with the advantage that LiDAR sensors work in all lighting and weather conditions, unlike RGB cameras. 

The goal of semantic segmentation is to assign a semantic label to each atomic unit from the data being processed. For example, for 2D images, we assign labels to individual pixels, and in 3D point clouds, we assign a label to each individual point, enabling the identification of objects and their boundaries in the 3D space. By accurately segmenting LiDAR point clouds, autonomous vehicles can better perceive their surroundings and make more informed driving decisions. In order to create algorithms to learn how to perceive semantic classes within a point cloud, we require labeled data so the algorithms can learn from it. But collecting and labeling real-world data can be a costly and time-consuming process. For example, an already-dated Velodyne HDL-64E LiDAR scan can capture roughly 2,200,000 points in one second. Manually labeling these scans for a task such as semantic segmentation requires a lot of human effort. In contrast, if we use simulation software to generate synthetic LiDAR scans, we can eliminate the need for manual labeling.


There are already different real-world point-cloud semantic segmentation datasets, such as nuScenes \cite{caesar_nuscenes_2020}, SemanticKITTI \cite{behley_semantickitti_2019} or SemanticPOSS \cite{pan_semanticposs_2020}, but they have some limitations and disadvantages, such as limited variability, annotation errors (especially on boundaries of colliding objects) or limited class coverage, which can introduce biases on the models trained using them.





Synthetic datasets offer many advantages compared to real datasets, such as reducing the time and cost associated with data collection and labeling, while ensuring a perfect and consistent labeling approach, and removing human error from the annotation process. By creating a synthetic dataset, it is also easier to adjust the variability of the data, for example by modifying the environment, or the elements from each class, also synthetic data allows us to simulate edge cases or extreme scenarios that may be difficult to capture in real-world data. 

But synthetic datasets also have their disadvantages, as they may not be representative of the real-world scenarios an autonomous vehicle may find. If we only have samples of one type of car in our synthetic dataset, such as sports cars, an algorithm that learned from that synthetic data could have issues identifying other types of cars, such as SUVs. 


Also, the quality of synthetic datasets also relies heavily on the accuracy of the simulation software used to generate them, which may introduce biases or inaccuracies, and may not capture the full range of variability and complexity present in real-world LiDAR data. For example, 3D objects used in video-games and simulation software are built using polygons, they tend to have smooth surfaces, on which synthetic LIDAR sensors produce a perfect surface, but in the real world, LiDAR sensors are noisy, and surfaces are rarely perfectly flat. But the simulation of these imperfect surfaces would require additional design efforts and computational capabilities, up to a point where it may become untractable. 

Therefore, it is crucial to find adequate approaches to create and use synthetic datasets to train LiDAR semantic segmentation models that can perform better than models trained on real data alone.

In this work, we present two main different contributions:
\begin{itemize}
    \item A novel synthetic dataset for semantic segmentation on LiDAR, designed specifically to be similar to SemanticKITTI, with a class distribution adjusted for realism, and more labeled data. We prove its usefulness by improving the performance of two different algorithms from the state of the art.
    \item An adapted version of the CARLA simulator with more specific labels for critical objects such as different types of vehicles, so it can be used to generate LiDAR scans that are better aligned with real datasets, such as SemanticKITTI or nuScenes \cite{caesar_nuscenes_2020} and tools for class distribution adjustment. 
\end{itemize}

\section{Related Work}
In this section, we introduce semantic segmentation on LiDAR imaging, presenting different common strategies for this task, and then we introduce existing datasets for semantic segmentation on LiDAR imaging.

\subsection{Semantic Segmentation on LiDAR imaging.}

Semantic segmentation is a technique that consists of classifying data at an atomic level, for example, classifying individual pixels in an image, compared to only classifying an image into a class. As such, semantic segmentation also preserves location, providing us with more information than just classifying the whole image or scene. It has multiple applications: with 2D images, it can be used for aerial imaging, autonomous driving, and medical applications among others. In this work, we apply semantic segmentation to LiDAR scans, which are 3D images of the world surrounding a LiDAR sensor, and in general, they are represented as a 3D Point cloud. With LiDAR scans, semantic segmentation can also be used for autonomous driving or aerial imaging, and on other tasks such as robotic navigation, augmented reality, or scene completion. 

There is a big difference between image and point cloud segmentation, while pixels on an image share a structure, and there are no empty spaces between pixels in the image representation, so they can be processed using 2d-convolutional layers. 3D point cloud segmentation is a more challenging task, as there is a lot of data sparsity (empty space between points) and points are not distributed in the space following a constant distribution. Also, there are other issues: while a 2D image is \textit{discrete}, as it has evenly distributed pixels along a grid, the 3D point clouds are \textit{continuous}, and therefore, they can not be processed with convolutions without some pre-processing, unlike 2d images. There are four main approaches to 3D point cloud segmentation: in \textit{point-based} methods, the data is processed as a point-cloud vector. First point-cloud segmentation models were built using the points as input \cite{charles_pointnet_2017}\cite{qi_pointnet_2017}\cite{wang_dynamic_2019}, although the trend shifted to other input methods to better exploit tools such as convolutional networks, and offering better results than previous point-based methods, up until WaffleIron \cite{puy_using_2023}, which offers state-of-the-art performance while being a point-based method. In \textit{projection-based} methods, the point clouds are converted to a 2d image using a projection, either cylindrical or spherical. The main advantage of this is allowing to process points as a flat image, transforming the task into a 2D segmentation process \cite{wu_squeezeseg_2018}\cite{wu_squeezesegv2_2019}\cite{xu_squeezesegv3_2021}. The main issue with these projections is the information loss upon projection. In RangeFormer \cite{kong2023rethinking} they introduce a new way to process these projections while reducing the loss of information, reaching state-of-the-art performance. \textit{Voxel-based} methods divide the 3D space of the point cloud into an even grid of 3D pixels, called voxels, and use this grid as input for the algorithms. This voxelization produces a grid with a lot of empty voxels, due to the sparsity of the point cloud. Thanks to Sparse Convolutions (SSCN) \cite{graham_3d_2018}, it became much more efficient to process these voxels, by reducing the number of required operations when processing empty voxels. As voxels discretize the continuous space of the point cloud, fine-grained information is lost in the process. For this reason, voxel-based methods usually rely on additional post-processing to perform a fine-grained, per-point segmentation \cite{tang_searching_2020}\cite{liu_point-voxel_2019}\cite{cheng_s3net_2021}. There are other voxelization strategies, for example, Cylinder3D\cite{zhu_cylindrical_2021} uses a cylindrical voxelization space, SphereFormer uses a spherical grid with radial voxels \cite{lai2023spherical}. At last, there are \textit{hybrid} methods that combine multiple methods in one pipeline, to alleviate weaknesses from each method by running other models that use different information in parallel, often exchanging feature information between models, and then fusing their predictions \cite{xu_rpvnet_2021}\cite{cheng_af_2021}. Other methods even include information from RGB cameras, such as 2DPASS \cite{yan_2dpass_2022}. As hybrid methods run multiple models in parallel, they are very demanding computationally, but in general, they obtain the best results in different state-of-the-art benchmarks.

The main disadvantage of these models is that as we run different types of inputs and models in parallel, they require even more memory and computing power to run, and train, also their architecture is more complex and harder to replicate, but in general, they obtain the best results in different benchmarks fron the state o the art.

\subsection{3D Semantic Segmentation Datasets}

%

The first datasets for 3D point cloud semantic segmentation were designed for different, non-urban, tasks. For example, ModelNet \cite{zhirong_wu_3d_2015} is a dataset designed for the classification and segmentation of different objects, and it provides 12,311 3D models of different objects from 40 categories. ScanNet \cite{dai_scannet_2017} provides 3D reconstructions of indoor scenes, with 2.5M views in 1513 scenes, annotated with 3D camera poses. As newer smart and autonomous cars often equip different LiDAR sensors and more powerful processing units, there is an increasing interest in segmenting urban point clouds, such as those from the KITTI Odometry \cite{geiger_are_2012} dataset. SemanticKITTI \cite{behley_semantickitti_2019} provides ground-truth labels for the LiDAR scans from KITTI Odometry, has 28 different semantic classes and over 20,000 labeled scans from 10 sequences captured on different locations, and it also provides a test server to evaluate algorithms on the other 11 sequences. Other urban datasets for LiDAR semantic segmentation have been published such as nuScenes \cite{caesar_nuscenes_2020} has 23 classes, and 40,000 scans, or Semantic Poss \cite{pan_semanticposs_2020} with 14 classes and 2,988 scans.

The problem with all these datasets is that they require vast amounts of time for both data collection and annotation. As such, synthetic datasets have started gaining attention, as they have been proven effective in training algorithms from other modalities, such as semantic segmentation in RGB images \cite{park_dat_2022}, and they can be generated using already existing software such as simulators or even video games. For example, in SqueezeSeg \cite{wu_squeezeseg_2018}, the authors create a simple synthetic LiDAR semantic segmentation dataset (3 classes only) using the GTA V video game and train their model using synthetic data and the LiDAR scans from the KITTI Odometry dataset \cite{geiger_are_2012}.

CARLA \cite{dosovitskiy_carla_2017} is a simulation tool that includes support for many different types of sensors, such as RGB, depth and semantic cameras, LiDAR, and Radar, among many others. It has support for 23 different semantic classes, 7 different maps, and multiple vehicle and person models. Still, it has issues, such as only one vehicle class, with no specific classes for different types of vehicles.

With the introduction of tools like CARLA, more people started publishing synthetic LiDAR datasets, such as SynLiDAR \cite{xiao_transfer_2022} with 32 classes and 200,000 scans or KITTI-CARLA \cite{deschaud_kitti-carla_2021}, with 23 classes and 35,000 scans. However, it is common for these datasets to be designed as standalone datasets, hence missing classes from real datasets, as KITTI-CARLA does, or without taking into account class distribution in real-world environments, as is the case with SynLiDAR. In Table \ref{tab:datasetcomparison}, we show a quick comparison between different sequential LiDAR datasets, both synthetic and real.

\setlength{\tabcolsep}{1pt}
\begin{table}[t]
    \centering
    \begin{tabular}{l|cccc}
    & & & & \footnotesize{Realistic Class} \\
    Dataset & Type &  \#Points &  \#Classes$^1$ & \footnotesize{Distribution}\\
    \hline
    SemanticKITTI &  Real & 4,549M & 28(19)  & \cmark \\
    nuScenes & Real &  1,400M & 32(8) &\cmark\\ 
    \small{SemanticPOSS} & Real  & 216M & 14(10) & \cmark\\
    \small{KITTI-CARLA} \cite{deschaud_kitti-carla_2021} & \small{Synthetic}  & 4,500M & 23(9) &\xmark\\
    SynLiDAR \cite{xiao_transfer_2022} & \small{Synthetic}  & 19,482M &  32(19)& \xmark\\
    \textit{Ours} & \small{Synthetic} & 7,200M & 30(15)   &\cmark \\
    \end{tabular}
    \caption{Dataset comparison between different real and synthetic datasets for LiDAR semantic segmentation. $^1$ the number between brackets means the amount of classes shared with SemanticKITTI validation classes.}
    \label{tab:datasetcomparison}
\end{table}

Synthetic datasets offer the primary advantage of reducing the time and effort required for data capture and labeling, as labels can be automatically generated and include both cloud point and semantic labels. However, they also have limitations, such as a domain gap between synthetic and real-world data, and the challenge of realistically reproducing real-world datasets can also be a time-consuming process. 

\section{Dataset Generation}

\subsection{Simulator Settings}

Using CARLA \cite{dosovitskiy_carla_2017} simulator, we have modified and adjusted the simulator to include additional classes and generated a synthetic dataset similar to SemanticKITTI.

We configured our LiDAR sensor parameters to be equal to those used by SemanticKITTI. In the public toolkit for the SemanticKITTI dataset, we found sensor parameters for vertical field-of-view (from -3 to 25 degrees), and the number of channels in the sensor (64). By analyzing the SemanticKITTI dataset, we observed that scans in the dataset have a maximum range consistently within 80$\pm$0.3 meters, and the mean amount of points in a scan was about 130,000, depending on the scene. Accordingly, we adjusted the simulator's parameters to match those values.

To choose the car that would carry the LiDAR sensor in the simulator, we had to take into account the vehicle type used to capture the SemanticKITTI scenes (seemingly, a Volkswagen Passat based on their images), to ensure the amount and location of points captured from the ego car, such as the car hood, is similar to those from SemanticKITTI.

The next issue was that Carla only has 10 classes, out of the 19 classes used in evaluation in SemanticKITTI. KITTI-CARLA \cite{deschaud_kitti-carla_2021} builds a synthetic dataset for semantic segmentation on LiDAR scans, using the default CARLA classes, and therefore it only has 9 out of 19 classes in common with SemanticKITTI, missing key classes, such as specific vehicle classes. We modified CARLA to fix this weakness, adding vehicle classes such as car, bicycle, motorcycle,  moving bicycle and motorcycle, bicycle and motorcycle rider and truck. We also fixed label matching between real and synthetic objects such as guardrails, labeled as other-object in CARLA while their real-world counterpart is labeled as a fence on SemanticKITTI.

With our modifications, we were able to include 15 out of the 19 classes from SemanticKITTI validation. We did not include the other-vehicle, other-ground, parking, and trunk classes, as significant simulator changes were necessary. 

\begin{figure}[t]
\begin{center}
   \includegraphics[width=\linewidth]{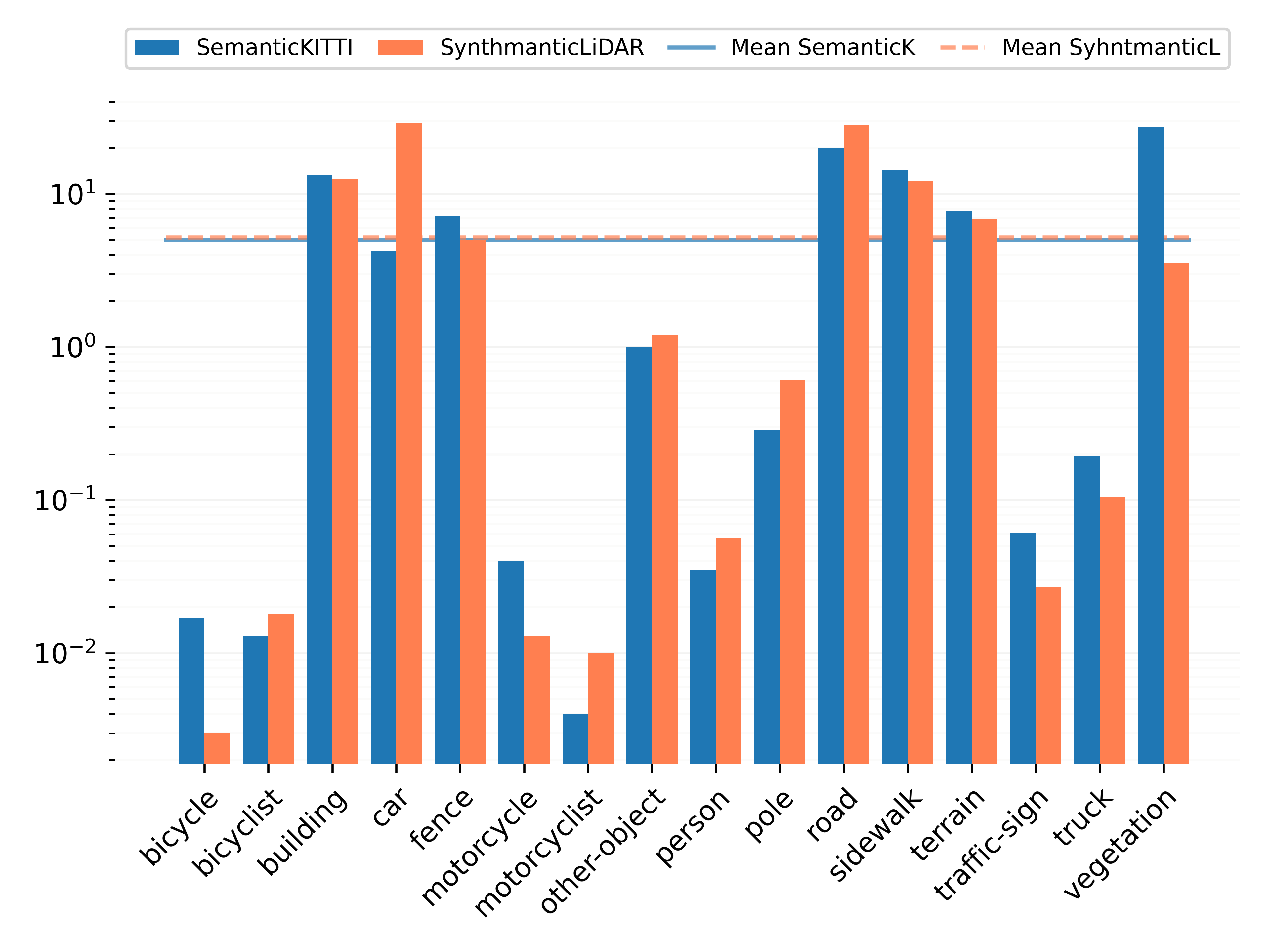}
\end{center}
   \caption{Proportion of labeled points in the dataset for classes shared between SemanticKITTI and our dataset, SynthmanticLiDAR. Logarithmic to visualize underrepresented classes.}
\label{fig:class_distribution}
\end{figure}

We also worked on being able to adjust the class distribution of the generated data. Other synthetic datasets, such as SynLiDAR \cite{xiao_transfer_2022}, provide a vast number of scans with many points for all the different classes, but in real data, this is not the case, as there are classes heavily underrepresented, such as bicyclist and motorcyclist. Therefore, we prepared our generation script to adjust to the real class distribution as much as possible and, as far as we know, we are the only synthetic LiDAR dataset that does this. We tested our generation method by creating our dataset following the class distribution from SemanticKITTI. In Figure \ref{fig:class_distribution} we compare the proportions of points from each class in the SemanticKITTI validation class list both in our dataset and in the SemanticKITTI dataset.


\subsection{Dataset Definition}
In total, our \textit{SynthmanticLiDAR} dataset is composed of 8 sequences of 6,000 scans each from 7 unique maps, for a total of 48,000 scans, which is more than double the number of labeled scans in the SemanticKITTI train subset. We also define a smaller subset, \textit{SynthmanticLiDAR-LT} composed from the first 2,000 scans from each sequence. In Figure~\ref{fig:examples} we show scans from three different sequences, and how that 3D LiDAR scan looks like in the simulator.

\section{Experiments}

\setlength{\tabcolsep}{2pt}
\begin{table*}[t]
    \centering
    \begin{tabular}{l|ccccccccccccccccccc|c}
    Method & \rot{car} & \rot{bicycle} & \rot{motorcycle} & \rot{truck} & \rot{\textit{other-vehicle}} & \rot{person} & \rot{bicyclist} & \rot{motorcyclist} & \rot{road} & \rot{\textit{parking}} & \rot{sidewalk} & \rot{\textit{other-ground}} & \rot{building} & \rot{fence} & \rot{vegetation} & \rot{\textit{trunk}} & \rot{terrain} & \rot{pole} & \rot{traffic-sign} & \rot{mIoU} \\
    \hline
    SPVCNN & 95.7 & \textbf{47.7} & \textbf{49.5} & 47.2 & 48.3 & \textbf{64.1} & 66.7 & 48.2 & 88.5 & 57.7 & 70.7 & 23.2 & 90.1 & 63.9 & \textbf{84.5} & \textbf{67.7} & \textbf{69.0} & 53.1 & 62.1 & 63.0 \\
    SPVCNN-F & \textbf{95.8} & \textbf{47.7} & 47.2 & \textbf{48.4} & \textbf{49.0} & 63.2 & \textbf{69.7} & \textcolor{cyan}{\textbf{49.0}} & \textbf{88.9} & \textbf{58.4} & \textbf{71.4} & \textcolor{cyan}{\textbf{24.0}} & 89.9 & 63.6 & 84.4 & 67.2 & 68.7 & \textbf{54.0} & \textbf{62.6} & \textcolor{cyan}{\textbf{63.3}} \\
    SPVCNN-LT & \textcolor{cyan}{\textbf{95.7}} & 45.6 & 44.5 & \textcolor{cyan}{\textbf{48.0}} & 47.6 & 62.6 & \textcolor{cyan}{\textbf{68.6}} & \textbf{59.1} & \textcolor{cyan}{\textbf{88.8}} & \textcolor{cyan}{\textbf{58.3}} & \textcolor{cyan}{\textbf{71.1}} & \textbf{26.8} & \textbf{90.3} & \textbf{64.7} & 84.2 & 66.6 & 68.1 & \textcolor{cyan}{\textbf{53.2}} & \textcolor{cyan}{\textbf{62.4}} & \textbf{63.5} \\
    \hline
    SSV3 & 81.4 & 16.0 & 25.3 & 3.7 & 13.3 & 34.0 &  33.1 & 13.5 &  \textbf{88.8} &  \textbf{52.8} & 68.4 &  \textbf{21.9} & 76.1 & 43.3 & 75.6 & 44.1 & 59.9 & 30.3 & 30.6 & 42.7 \\
    SSV3-F & \textcolor{cyan}{\textbf{84.2}} & \textbf{22.8}  & \textbf{28.8} & \textcolor{cyan}{\textbf{4.2}} & \textcolor{cyan}{\textbf{15.6}} & \textbf{38.2} & \textbf{33.4} & 9.0 & 88.1 & 51.2 & \textbf{68.9} & 21.8 & \textcolor{cyan}{\textbf{76.7}} & \textcolor{cyan}{\textbf{44.6}} & \textbf{76.6} & \textbf{44.9} & \textcolor{cyan}{\textbf{61.9}} & \textcolor{cyan}{\textbf{31.0}} & \textbf{35.3} & \textcolor{cyan}{\textbf{44.1}} \\
    SSV3-LT & \textbf{84.4} & \textcolor{cyan}{\textbf{20.7}}  & \textcolor{cyan}{\textbf{26.8}} & \textbf{6.2} & \textbf{17.1} & \textcolor{cyan}{\textbf{35.5}} & 32.4 & \textbf{19.7} & 87.9 & 52.2 & \textcolor{cyan}{\textbf{68.6}} & 19.9 & \textbf{77.5} & \textbf{45.2} & \textcolor{cyan}{\textbf{76.1}} & 42.0 & \textbf{62.6} & \textbf{31.7} & \textcolor{cyan}{\textbf{33.8}} & \textbf{44.2} \\
    \end{tabular}
    \caption{Test results after fine-tuning for SPVCNN and SqueezeSegV3 (SSV3) models, pre-trained on SynthmanticLiDAR and fine-tuned with SemanticKITTI. Results represent IoU (\%),\textit{-F} means it was pre-trained with the full SynthmanticLiDAR-F, and \textit{-LT} means it was pre-trained with SynthmanticLiDAR-LT. Results in \textbf{bold} mean best score and results in \textcolor{cyan}{\textbf{cyan}}  mean better than baseline.}
    \label{tab:algorithms_comparison}
\end{table*}

To show the usefulness of our dataset, we have evaluated the performance of two different algorithms for semantic segmentation in 3D point clouds before and after including both versions of our dataset in their training process. 

We have selected two segmentation algorithms: the projection-based SqueezeSegV3 \cite{xu_squeezesegv3_2021} and the voxel-based SPVCNN \cite{tang_searching_2020} models. This choice was made due to their open-source nature and public availability on GitHub and their favorable rankings on scoreboards in comparison to other projection-based and voxel-based algorithms. 

\begin{figure}[t]
\begin{center}
   \includegraphics[width=\linewidth]{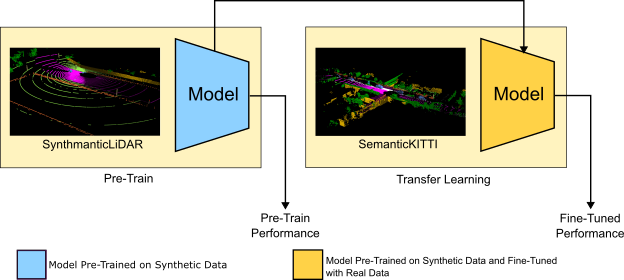}
\end{center}
   \caption{Scheme followed when training the point cloud semantic segmentation models. First, we pre-trained models using one of the two versions of our synthetic dataset, and then we fine-tuned them using the real data from SemanticKITTI.}
\label{fig:training_scheme}
\end{figure}

We trained both algorithms using a naive transfer-learning approach, following the schema shown in Figure~\ref{fig:training_scheme}. First, we pre-trained the models using either the full Synthmantic LiDAR dataset (referred to as Synthmantic LiDAR-F in this section) or the Synthmantic LiDAR-LT subset. Then, we fine-tune these models using the real SemanticKITTI dataset. We have also trained baselines with real data for comparison. In all training steps, we used the coordinates of the points in the point cloud as input for both algorithms.

Table~\ref{tab:algorithms_comparison} contains the baselines of both methods, and the results obtained after fine-tuning both variations of each architecture, the one pre-trained with SynthmanticLiDAR-F and the one pre-trained with SynthmanticLiDAR-LT. During the pre-training, we used 48,000 scans from 8 sequences for SynthmanticLiDAR-F and 16,000 scans from 8 sequences in SynthmanticLiDAR-LT. In the fine-tuning step, we used the standard training sequences from SemanticKITTI (sequences 00-07 and 09-10). In both training steps, sequence 08 from SemanticKITTI was used for validation epochs. All the hyperparameters can be found in the supplementary material.

For both models, we improve the overall performance by pre-training on either Synthmantic LiDAR-F or Synthmantic LiDAR-LT and then finetuning with the real SemanticKITTI. For SPVCNN, we improve in both cases in 12 out of 19 classes, and with SqueezeSegV3 we improve in 15 out of 19 classes with the Synthmantic LiDAR-F, and in 14 out of 19 classes with Synthmantic LiDAR-LT version. It is particularly noticeable how we improve in the most underrepresented class, motorcyclist, by a considerable amount (10.9\% for SPVCNN and 6.2\% for SqueezeSegV3) when pre-training with the SynthmanticLiDAR-LT subset, while this is not the case when pre-training with SynthmanticLiDAR-F, where performance even decreases for the SqueezeSegV3 model. We believe this behavior is due to the class being heavily underrepresented in the real data, with only 90,000 labeled points in the whole train set of SemanticKITTI, only 5 points on the validation set, and an unknown amount of points in the test set, which we assume to be similar in distribution to the number of points in the training set. In total, the number of points labeled as motorcyclist in SemanticKITTI train and validation sets add up to 0.004\% of the total labeled points, with an average per class of 6\% (horizontal line in Figure~\ref{fig:class_distribution}). This makes any small classification errors have a huge impact on the IoU metric for the motorcyclist class.

\setlength{\tabcolsep}{2pt}
\begin{table*}[h]
    \centering
    \begin{tabular}{l|ccccccccccccccc|c}
    Method & \rot{car} & \rot{bicycle} & \rot{motorcycle} & \rot{truck} & \rot{person} & \rot{bicyclist} & \rot{motorcyclist} & \rot{road} & \rot{sidewalk} & \rot{building} & \rot{fence} & \rot{vegetation} & \rot{terrain} & \rot{pole} & \rot{traffic-sign} & \rot{mIoU} \\
    \hline
    SPVCNN-F & 69.7 &3.8 & 4.9 & 4.8 & 23.6 & 27.3& 2.9 & 11.8 & 23.4 & 34.0 & 25.1 & 48.0 & 9.2 & 22.7 & 5.1 & 21.1 \\
    SPVCNN-LT & 74.7 & 0.4 & 5.2 & 7.0 & 21.5 & 15.7 & 6.9 & 12.7 & 24.8 & 18.2 & 22.0 & 40.9 & 18.4 & 14.3 & 4.8 & 19.2 \\
    \hline
    SSV3-F & 4.2 & 1.4 & 1.1 & 0.2 & 5.4 & 0.7 & 0.1 & 7.3 & 21.8 & 23.8 & 11.7 & 31.2 & 14.8 & 11.2 & 7.9  & 9.52 \\
    SSV3-LT &  12.8 & 0.4 & 0.9 & 1.4 & 2.5 & 0.4 & 23.8 & 16.6 & 32.6 & 10.9 & 20.0 & 15.1 & 7.5 & 3.8 & 7.8 & 9.91 \\
    \end{tabular}
    \caption{Results for SPVCNN and SqueezeSegV3 (SSV3) models, pre-trained on SynthmanticLiDAR, evaluated on SemanticKITTI. Results represent IoU (\%), \textit{-F} means it was pre-trained with the full SynthmanticLiDAR dataset, and \textit{-LT} means it was pre-trained with SynthmanticLiDAR-LT.}
    \label{tab:algorithms_comparison_pre}
\end{table*}

\begin{figure}[t]
\begin{center}
   \includegraphics[width=\linewidth]{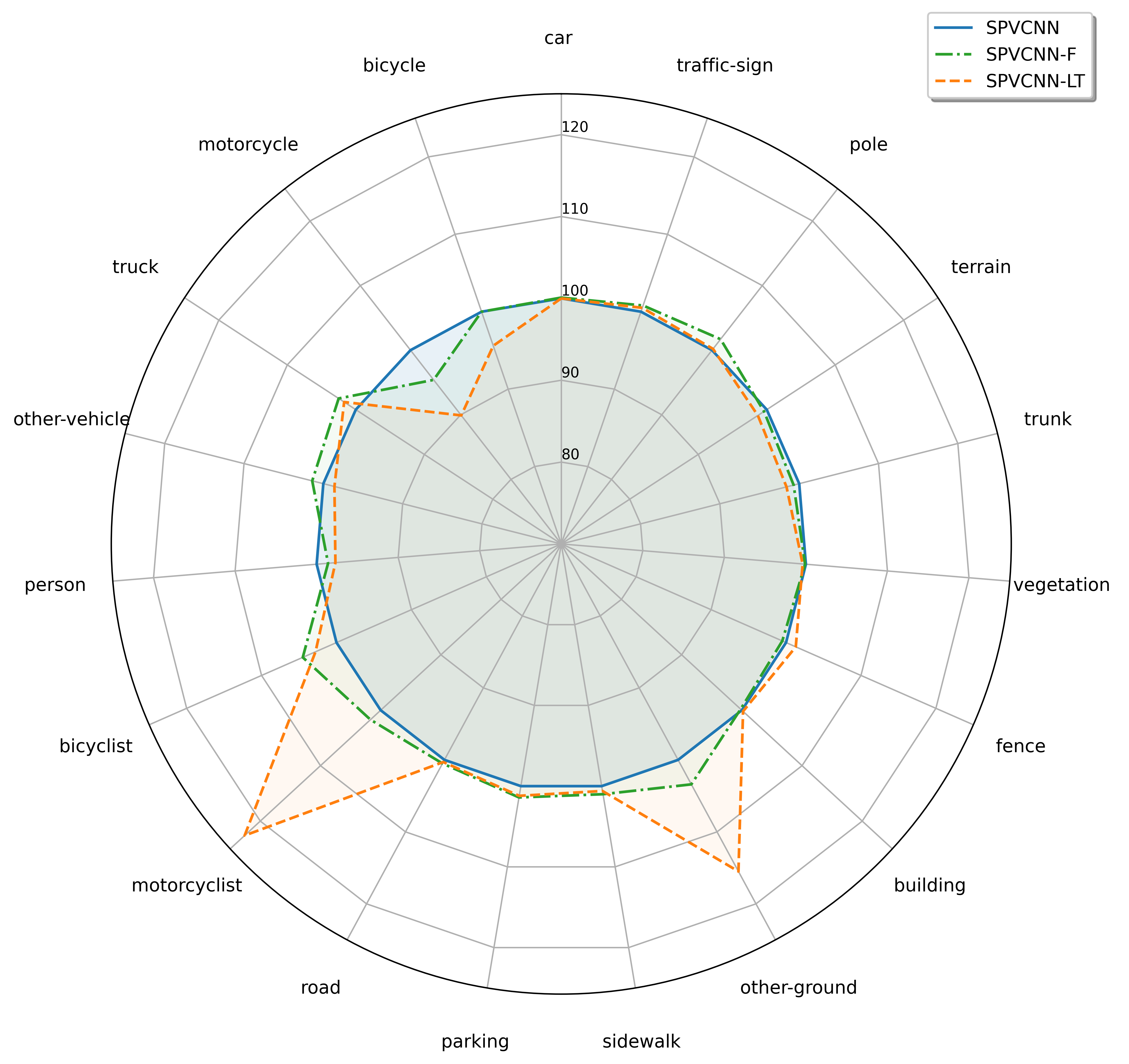}
\end{center}
   \caption{IoU scores for different versions of the SPVCNN algorithm, represented as a percentage increment over the baseline model.}
\label{fig:IoUs_spvcnn}
\end{figure}

\begin{figure}[t]
\begin{center}
   \includegraphics[width=\linewidth]{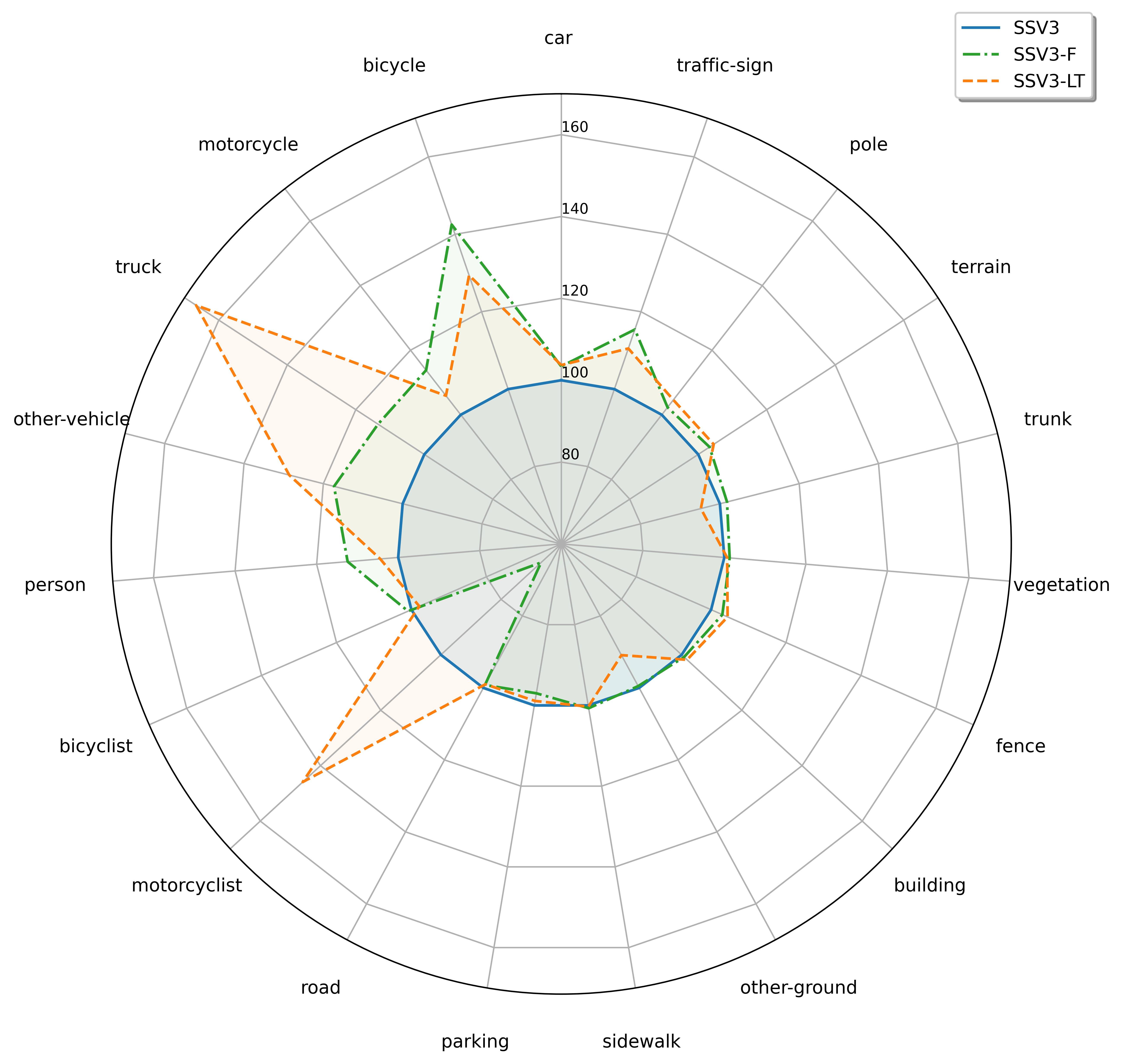}
\end{center}
   \caption{IoU scores for different versions of the SqueezeSegV3 algorithm represented as a percentage increment over the baseline model. Small increments result in a large percentage increase in low-performance classes.}
\label{fig:IoUs_ssv3}
\end{figure}

Although in Table~\ref{tab:algorithms_comparison} it can be observed that the model pre-trained with SynthmanticLiDAR-LT reaches higher mIoU for both methods, in Figures~\ref{fig:IoUs_spvcnn}~and~\ref{fig:IoUs_ssv3} we show that this score is mainly influenced by obtaining high performance in two classes, one of them being the class motorcyclist, while it consistently scores worse than the models trained with SynthmanticLiDAR-F on other more represented classes, such as bicycle or motorcycle. This suggests that, at least with our approach, the small dataset can help with some classes and the expense of worse performance in others, while the full dataset offers a more balanced performance.

To measure how aligned is SynthmanticLiDAR with the real SemanticKITTI dataset, we evaluate both models after pre-training, without finetuning.

Table~\ref{tab:algorithms_comparison_pre} shows the results obtained by evaluating the models trained only on synthetic data on the test set from SemanticKITTI. The SPVCNN model trained with Synthmantic LiDAR-F outperforms the model trained on the SynthmanticLiDAR-LT subset. But, after finetuning on real data, this difference in performance seems to be lost, as the model trained with SynthmanticLIDAR-F yields a slightly worse mIoU than the one pre-trained on SynthmanticLiDAR-LT. SqueezeSegV3 has a similar behavior but obtains a very low IoU without finetuning for both datasets. This could be due to the projection of synthetic point clouds presenting more dissimilarities with the real projected point clouds.

\section{Conclusions}
In this paper, we have presented a novel synthetic dataset for semantic segmentation on LiDAR scans, SynthmanticLiDAR, designed specifically to be used with SemanticKITTI, \\
\vspace{-5pt} \\ and with our experiments, we have proved its capabilities to improve the performance of algorithms using a naive transfer-learning approach such as pre-training on SynthmanticLiDAR data and then finetuning to SemanticKITTI. By proving its usefulness with such a naive approach, we pave the way for future improvements in performance with more refined transfer-learning approaches. Moreover, our modified CARLA simulator has also been proven useful to generate sequences that can be used in conjunction with real datasets.

\bibliographystyle{IEEEbib}

\let\oldbibliography\thebibliography
\renewcommand{\thebibliography}[1]{
  \oldbibliography{#1}
  \setlength{\itemsep}{0pt} 
  \setlength{\parskip}{0pt} 
}

\bibliography{egbib}

\end{document}